%% file: main.tex
\documentclass[9pt,conference]{IEEEtran}
\usepackage[T1]{fontenc}
\usepackage{cite}
\usepackage{amsmath,amssymb}
\usepackage{algorithm}
\usepackage{algpseudocode}
\usepackage{booktabs,array,graphicx}
\usepackage{tabularx}
\usepackage[table]{xcolor}
\usepackage{pifont,makecell}
\usepackage{tikz}
\usetikzlibrary{arrows.meta,positioning,calc}
\usepackage[most]{tcolorbox}
\usepackage{url}
\usepackage[hidelinks]{hyperref}
\usepackage{balance}
\usepackage{multirow}
\usepackage{microtype}

\definecolor{cCrim}{HTML}{971B2F}
\definecolor{cInk}{HTML}{202429}
\definecolor{cGray}{HTML}{65717B}
\definecolor{cPale}{HTML}{F5F6F7}
\input{styles}

\newcommand{\thd}[1]{\textcolor{white}{\textbf{#1}}}
\newcommand{\tablecompact}{\setlength{\tabcolsep}{3.5pt}\renewcommand{\arraystretch}{0.9}}

\AtBeginDocument{%
 \setlength{\abovedisplayskip}{4pt plus 1pt minus 1pt}%
 \setlength{\belowdisplayskip}{4pt plus 1pt minus 1pt}%
 \setlength{\abovedisplayshortskip}{2pt plus 1pt}%
 \setlength{\belowdisplayshortskip}{2pt plus 1pt}}
\makeatletter
\renewcommand\section{\@startsection{section}{1}{\z@}
 {2.0ex plus .6ex minus .2ex}{0.5ex plus .2ex}
 {\normalfont\normalsize\centering\scshape}}
\renewcommand\subsection{\@startsection{subsection}{2}{\z@}
 {1.4ex plus .5ex minus .2ex}{0.4ex plus .2ex}
 {\normalfont\normalsize\itshape}}
\makeatother

\newcommand{\system}{\Reforge}
\newcommand{\V}{\textsc{verified}}
\newcommand{\B}{\textsc{bounded}}
\newcommand{\U}{\textsc{unchecked}}
\newcommand{\F}{\textsc{failed}}
\newcommand{\sm}[1]{s_{\mathrm{#1}}}
\newcommand{\eqnumcirc}[1]{%
 \tikz[baseline=(n.base)]{\node[circle,draw=cInk,fill=black!12,
  line width=0.45pt,inner sep=0.7pt,minimum size=1.15em,
  font=\footnotesize] (n) {#1};}}
\makeatletter
\renewcommand{\tagform@}[1]{\maketag@@@{\eqnumcirc{#1}}}
\makeatother
\newtcolorbox{databox}[1]{colback=white,colframe=cGray,boxrule=0.6pt,
 arc=1mm,left=5pt,right=5pt,top=3pt,bottom=3pt,before skip=4pt,after skip=4pt,
 title={#1},fonttitle=\bfseries,fontupper=\footnotesize,
 colbacktitle=cPale,coltitle=cInk}
\DeclareMathOperator{\cossim}{cos}

\begin{document}
\title{REFORGE: LLM-Assisted RTL Bug Repair with\\
Structured Context and Verification Feedback}
\author{
\IEEEauthorblockN{Tanvir Hossain\textsuperscript{*}, Sudipta Paria\textsuperscript{\dag}, Aidan Gleason Myers\textsuperscript{\dag}, Swarup Bhunia\textsuperscript{\dag}, and Tamzidul Hoque\textsuperscript{*}}
\IEEEauthorblockA{tanvir@ku.edu, sudiptaparia@ufl.edu, myersaidan@ufl.edu, swarup@ece.ufl.edu, hoque@ku.edu\\
\textsuperscript{*}Dept.\ of EECS, University of Kansas, Lawrence, KS, USA, and \textsuperscript{\dag}Dept.\ of ECE, University of Florida, Gainesville, FL, USA}
}
\maketitle
\begin{tikzpicture}[remember picture,overlay]
\draw[black,line width=0.4pt] ([xshift=-0.5\textwidth,yshift=0.62in]current page.south) -- ++(\textwidth,0);
\node[text=blue,font=\scriptsize,anchor=north] at ([yshift=0.58in]current page.south) {This paper has been accepted for publication at DATE 2027. This is an initial author version. The final paper will appear in the conference proceedings.};
\end{tikzpicture}

\begin{abstract}
Large language models (LLMs) can write and repair RTL, but successful compilation or testbench execution does not establish defect removal, behavioral preservation, security, or implementation feasibility. Existing LLM-based RTL repair methods often report a single repair rate without retaining separate evidence for these properties, limiting reproducible comparison and integration into hardware design flows. We present \system{}, which, to our knowledge, is the first LLM-assisted RTL repair framework that maintains and accumulates structured repair evidence across iterations, allowing it to generate fixes that satisfy multiple objectives simultaneously, including functionality, behavioral equivalence, security, and cost efficiency. \system{} profiles buggy RTL, identifies structural and security-relevant features, localizes failing behavior, and uses verification-guided diagnosis to generate complete RTL repair variants. Each variant undergoes syntax checking, simulation, formal analysis, off-trigger preservation, security re-scan, and synthesis-based power, performance, and area analysis. The framework records verdicts, proof depths, counterexamples, tool availability, and cost measurements, then selects verified variants using source-, netlist-, activity-, and PPA-level similarity. We implement \system{} as a backend-independent repair-and-verification flow and evaluate it in 288 runs across three LLM backends and three benchmark suites. On CirFix, the proposed context and diagnosis increase runs delivering a checked repair from 32/63 to 51/63, while every configuration repairs the Hack\@DAC OpenTitan defect. Evidence records also show that a testbench-only criterion reports 3.9 times more repairs than the full set of applicable checks, security-clean outputs may lack functional verification, and different RTL repairs can implement equivalent hardware. These results show that LLM-generated RTL repairs require explicit multi-stage verification evidence rather than a single pass/fail outcome.

\end{abstract}


\input{sections/intro}
\input{sections/background}

\input{sections/method}

\input{sections/setup}
\input{sections/results}

\balance
\input{sections/closing}
\clearpage
\bibliographystyle{IEEEtran}
\bibliography{references}

\end{document}

%% file: styles.tex
\definecolor{tagCodex}{RGB}{45,124,110}
\definecolor{tagClaude}{RGB}{193,110,72}
\definecolor{tagGrok}{RGB}{72,92,145}

\providecommand{\xspace}{\space}
\DeclareRobustCommand{\ptag}[2]{%
  \tikz[baseline=(X.base)]{%
    \node[
      draw=#1!70,
      fill=#1!12,
      line width=0.4pt,
      rounded corners=2.2pt,
      inner xsep=3.8pt,
      inner ysep=1.6pt,
      font=\sffamily\bfseries\scriptsize
    ] (X) {\textcolor{#1}{#2}};
  }%
}

\DeclareRobustCommand{\Reforge}{\ptag{cCrim}{REFORGE}\xspace}

\DeclareRobustCommand{\Codex}{\ptag{tagCodex}{Codex}\xspace}
\DeclareRobustCommand{\Claude}{\ptag{tagClaude}{Claude}\xspace}
\DeclareRobustCommand{\Grok}{\ptag{tagGrok}{Grok}\xspace}
\DeclareRobustCommand{\Sonnet}{\ptag{tagClaude}{Sonnet}\xspace}

\definecolor{tagRQGray}{RGB}{225,225,227}
\DeclareRobustCommand{\rqtag}[1]{%
  \tikz[baseline=(X.base)]{%
    \node[
      draw=black,
      fill=tagRQGray,
      line width=0.4pt,
      rounded corners=2.2pt,
      inner xsep=3.8pt,
      inner ysep=1.6pt,
      font=\sffamily\bfseries\scriptsize
    ] (X) {\textcolor{black}{#1}};
  }%
}
\DeclareRobustCommand{\RQOne}{\rqtag{RQ1}\xspace}
\DeclareRobustCommand{\RQTwo}{\rqtag{RQ2}\xspace}
\DeclareRobustCommand{\RQThree}{\rqtag{RQ3}\xspace}

%% file: sections/intro.tex
\vspace{-1em}
\section{Introduction}
\label{sec:intro}
Large language models can generate RTL modules, complete
testbenches, and explain unfamiliar hardware. These capabilities shorten
the code development time, but do not reduce the verification
burden, which consumes roughly half of an ASIC
project~\cite{foster2022wilson}. LLM-generated RTL can contain
 many bugs that update a register under the wrong condition, violate a handshake, or
leave a security lock writable~\cite{mitre2025mihw}. Consequently, any design
flow that accepts machine-generated RTL also needs a
systematic repair-and-verification process. Previous work, that address this critical need, uses LLMs to generate complete edits for syntax errors and hardware
weaknesses~\cite{tsai2024rtlfixer,ahmad2024hsbugfix,llm_survey}.
We define an \emph{RTL repair variant} as a complete RTL
modification for a specific bug. Generation is only one part
of bug-free RTL repair. Each variant must
also be checked for bug removal, preservation of previously
correct behavior, avoidance of masked or newly introduced weaknesses, and
implementation cost.

Existing evaluations often report a repair rate, the fraction
of repair attempts that pass a selected test or verification check
(Table~\ref{tab:positioning}). This single number can hide important differences
in evidence. For example, a variant may pass simulation while a
sequential corner case remains unproved. A preservation check may not exercise 
relevant behavior. PPA cannot be compared without a valid reference
~\cite{qi2015plausibility,long2016search,smith2015cure}. The problem is therefore
not only to generate a repair, but also to retain the evidence from each check.
A typed record can route a bounded result to deeper verification, reject a
regression, or prevent an unchecked result from being reported as success.

Repeated generation can produce different RTL descriptions that synthesize to
functionally equivalent hardware. Although one verified repair
can correct a bug, alternative repairs can help a designer choose a cleaner RTL or
a better PPA result. Implementation diversity, therefore, refers to differences
in the synthesized structure or PPA, not the source text alone. Repair, evidence
strength, and implementation diversity must be measured separately.

These observations motivate three questions. \RQOne{}: How do structural design
context and verification feedback affect LLM-based generation of RTL repair
variants? \RQTwo{}: What evidence does each design check
provide for an RTL repair variant? \RQThree{}: Do verified RTL
repair variants remain different after synthesis and what computational and
implementation costs are associated with generating them?

\system{} answers these questions by separating proposal, acceptance, and
selection. An LLM proposes variants; simulation, formal, security, and synthesis
tools record a separate result for each design check; and a
selector compares verified variants at the source, netlist,
switching-activity, and PPA levels. Three prompt configurations isolate the
available evidence: \emph{bare} provides only source and a repair request,
\emph{unguided} adds structural context and transformation directions, and
\emph{guided} also returns the prior round's blocking diagnosis.

This paper makes following major contributions: \noindent\eqnumcirc{1}\enspace\textbf{Structured variant generation.}
We build an iterative flow that generates several repair variants using design
structure, transformation choices, and feedback from failed checks. \noindent\eqnumcirc{2}\enspace\textbf{Checks across the design flow.}
Each variant passes through syntax, simulation, formal, and
security checks before it can be reported as a supported repair.
\noindent\eqnumcirc{3}\enspace\textbf{Evaluation on three repair suites.}
We create a hardware-CWE repair suite and evaluate it alongside CirFix and an
external Hack@DAC OpenTitan bug. \noindent\eqnumcirc{4}\enspace\textbf{Complete repair tool flow.}
The framework connects open-source and commercial design tools and includes
area, timing, and power constraints when implementation evidence is available.
Together, these contributions provide a practical path from generated repair
variants to checked RTL suitable for research and design use.

The remainder of the paper positions the work in Section~II, develops the
repair framework in Section~III, describes the evaluation in Section~IV,
presents results in Section~V, and concludes with implications and limits in
Sections~VI--VII.
\vspace{-.7em}

%% file: sections/background.tex
\section{Related Work and Research Scope}
\label{sec:background}

\newcommand{\cY}{\ding{51}}
\newcommand{\cN}{\ding{55}}
\newcommand{\cP}{\ding{109}}
\newcommand{\cNA}{--}
\newcommand{\cU}{?}
\definecolor{hdr}{HTML}{971B2F}
\definecolor{hdr2}{HTML}{E8000D}
\definecolor{grp}{HTML}{D2D7DC}
\definecolor{alt}{HTML}{EDF1F4}
\newcommand{\hd}[1]{\textcolor{white}{\textbf{#1}}}
\newcommand{\vhd}[2]{\rotatebox[origin=c]{90}{\fontsize{6}{6.2}\selectfont
  \shortstack[c]{\textcolor{white}{\textbf{#1}}\\[-2pt]\textcolor{white}{\textbf{#2}}}}}
\newcommand{\vhs}[1]{\rotatebox[origin=c]{90}{\fontsize{6}{6.2}\selectfont
  \textcolor{white}{\textbf{#1}}}}
\begin{table}[t]
\centering
\caption{Supported acceptance evidence in prior RTL work and this paper. \cY{} full,
\cP{} partial, \cN{} absent, \cNA{} not applicable.}
\label{tab:positioning}
\footnotesize
\setlength{\tabcolsep}{0.75pt}
\renewcommand{\arraystretch}{0.84}
\begin{tabularx}{\columnwidth}{|>{\raggedright\arraybackslash}X|c|c|*{10}{c|}}
\hline
\rowcolor{hdr}
\textcolor{white}{\textbf{System (year)}} & \vhs{Task} & \vhd{Functional}{oracle} &
\vhd{Bug}{removal} & \vhd{Behavior}{preservation} & \vhd{Formal}{depth} &
\vhd{Counterexample}{feedback} & \vhd{Security}{check} & \vhd{Security}{re-scan} &
\vhd{PPA}{measurement} & \vhd{PPA}{constraint} & \vhd{Diverse}{selection} & \vhd{Context}{reuse} \\
\hline
\rowcolor{grp}\multicolumn{13}{|l|}{\textit{Functional RTL repair, localization, and debugging}} \\
\hline
CirFix (2022)~\cite{ahmad2022cirfix} & R & S+M & \cN & \cN & \cN & \cN & \cN & \cN & \cN & \cN & \cN & \cN \\
\rowcolor{alt}
RTL-Repair (2024)~\cite{laeufer2024rtlrepair} & R & S+F$_k$ & \cP & \cP & \cP & \cN & \cN & \cN & \cN & \cN & \cN & \cN \\
RTLFixer (2024)~\cite{tsai2024rtlfixer} & R$_{\text{syn}}$ & S & \cN & \cN & \cN & \cP & \cN & \cN & \cN & \cN & \cN & \cN \\
\rowcolor{alt}
MEIC (2024)~\cite{xu2024meic} & R & S+A & \cN & \cN & \cN & \cP & \cN & \cN & \cN & \cN & \cN & \cN \\
R3A (2025)~\cite{r3a2025} & R & S & \cN & \cN & \cN & \cP & \cN & \cN & \cN & \cN & \cN & \cN \\
\rowcolor{alt}
Location-is-Key (2025)~\cite{lik2025} & Lo & S & \cN & \cN & \cN & \cN & \cN & \cN & \cN & \cN & \cN & \cN \\
Fixbench-RTL (2025)~\cite{fixbench2025} & B & S & \cN & \cN & \cN & \cN & \cP & \cN & \cN & \cN & \cN & \cN \\
\rowcolor{alt}
BugGen (2025) \cite{buggen} & I & S & \cNA & \cNA & \cNA & \cP & \cN & \cN & \cN & \cN & \cN & \cN \\
\hline
\rowcolor{grp}\multicolumn{13}{|l|}{\textit{Quality-aware generation and optimization}} \\
\hline
AutoVeriFix (2025) \cite{autoverifix} & G & S & \cN & \cNA & \cN & \cP & \cN & \cN & \cN & \cN & \cN & \cN \\
\rowcolor{alt}
ChipSeek (2025)~\cite{chipseek2025} & G & S & \cN & \cNA & \cN & \cP & \cN & \cN & \cY & \cN & \cN & \cP \\
LLM-VeriPPA (2025)~\cite{thorat2025veripppa} & G & S & \cN & \cNA & \cN & \cP & \cN & \cN & \cY & \cN & \cN & \cN \\
\rowcolor{alt}
POET (2026)~\cite{yao2026poet} & O & S+F & \cNA & \cY & \cN & \cP & \cN & \cN & \cY & \cN & \cP & \cN \\
VeriRAG (2025)~\cite{verirag2025} & R$_{\text{DFT}}$ & F & \cP & \cY & \cN & \cP & \cN & \cN & \cN & \cN & \cN & \cN \\
\hline
\rowcolor{grp}\multicolumn{13}{|l|}{\textit{Hardware security: repair and detection}} \\
\hline
\rowcolor{alt}
CWEAT / HS-Fix (2022,'24)~\cite{ahmad2022cweat,ahmad2024hsbugfix} & R$_{\text{sec}}$ & S & \cN & \cN & \cN & \cN & \cY & \cP & \cN & \cN & \cN & \cN \\
Self-HWDebug (2024) \cite{selfhwdebug} & R$_{\text{sec}}$ & A & \cN & \cN & \cN & \cN & \cP & \cN & \cN & \cN & \cN & \cN \\
\rowcolor{alt}
SV-LLM (2025), Agentic Ver. (2026) \cite{svllm},\cite{agentic_vts} & V & S & \cN & \cNA & \cN & \cN & \cP & \cN & \cN & \cN & \cN & \cN \\
BugWhisperer, LASHED, MARVEL(2025)~\cite{bugwhisper,lashed2025,marvel2025} & D & \cNA & \cNA & \cNA & \cNA & \cNA & \cY & \cNA & \cNA & \cNA & \cNA & \cN \\
\hline
\rowcolor{hdr2}
\textcolor{white}{\textbf{This paper}} & \textcolor{white}{\textbf{R$_{\text{fn+sec}}$}} & \textcolor{white}{\textbf{S+F}} & \textcolor{white}{\cY} & \textcolor{white}{\cY} & \textcolor{white}{\cY} & \textcolor{white}{\cY} & \textcolor{white}{\cY} & \textcolor{white}{\cY} & \textcolor{white}{\cY} & \textcolor{white}{\cY} & \textcolor{white}{\cY} & \textcolor{white}{\cP} \\
\hline
\end{tabularx}
\begin{minipage}{\columnwidth}\scriptsize
\setlength{\parindent}{0pt}
\textbf{Task:} R functional repair, R$_{\text{syn}}$ syntax repair, R$_{\text{sec}}$ security repair, R$_{\text{DFT}}$ testability repair, G RTL generation, O PPA optimization, D bug detection, Lo bug localization, I bug injection, B benchmark, V verification workflow.
\textbf{Functional oracle:} S simulation, S+M simulation with manual review, S+A simulation with runtime assertions, A assertions, F formal equivalence to the \emph{original} design, F$_k$ bounded formal analysis of the failing trace. Remaining columns are defined below.
\end{minipage}
\vspace{-3pt}
\end{table}

This section positions \system{} against prior RTL repair, optimization, and
security-analysis work, with emphasis on the evidence used to accept a repair.

\vspace{-1em}
\subsection{Acceptance Evidence in Prior RTL Repair Systems}
\label{sec:bg-repair}

Prior works differ in both repair generation and acceptance
evidence. Table~\ref{tab:positioning} first identifies each
task and functional oracle, then compares bug removal, behavioral preservation,
formal depth and feedback, security analysis, PPA evaluation, diverse selection,
and context reuse. Diversity denotes verified alternatives that remain distinct
after synthesis, allowing designers to compare structural or PPA tradeoffs.

The first group comprises CirFix, RTL-Repair, RTLFixer, MEIC,
R3A, Location-is-Key, Fixbench-RTL, and BugGen. Across this group, methods
search for edits that pass the supplied testbench, use symbolic
synthesis, or iterate with compiler feedback, debugging, and retrieval
~\cite{ahmad2022cirfix,laeufer2024rtlrepair,tsai2024rtlfixer,xu2024meic,
r3a2025}. Contextual prompting has also been studied
separately~\cite{elnaggar2025context}. These studies demonstrate LLM-assisted
RTL repair under their respective oracles, but do not jointly report
preservation of correct behavior outside the bug-triggering
condition and formal proof depth. Because CirFix and RTL-Repair
use different oracles, bare prompting serves only as an internal baseline. Also, CirFix verify the outcomes with only a specific testbench and manual inspections.

The second group adds implementation evidence. ChipSeek,
LLM-VeriPPA, and POET
report implementation cost~\cite{chipseek2025,thorat2025veripppa,yao2026poet},
while POET also preserves the original design behavior.
Security studies cover another subset of columns: CWEAT,
HS-Fix, and Self-HWDebug address repair, whereas SV-LLM, Agentic Verifier,
BugWhisperer, LASHED, and MARVEL emphasize verification or detection
~\cite{ahmad2022cweat,ahmad2024hsbugfix,selfhwdebug,
svllm,agentic_vts,bugwhisper,lashed2025,marvel2025}.
Thus, relevant checks exist across prior systems, but no listed
system combines functional preservation, security improvement, implementation
cost, and diverse selection when accepting a repair.

\subsection{Scope and Terminology}\label{sec:bg-security}

\system{} addresses this gap through one verification process
that runs the applicable checks and records each result and formal proof depth.
The objective is not an unmatched repair-rate comparison, but a
traceable basis for every acceptance decision. For each selected
candidate, the process records bug removal, preservation depth, security changes,
measured cost, and its relation to other candidates. A candidate is not
evidence-complete while a required check remains unavailable.

Table~\ref{tab:positioning} summarizes system support; it does
not guarantee that every candidate reaches the strongest available check. For
evaluation, we use published CirFix and Hack@DAC benchmarks.
For security-related evaluations, we also manually construct CWE-buggy RTL
designs to create a benchmark for security bugs and their repair using our proposed framework. A differential security re-scan compares findings before and
after repair. Removal means that a finding disappears; relocation moves the same
weakness to another signal or location, while masking hides it from the detector
without removing it. Cost measurement reports area, delay, and power;
a cost gate also requires a valid reference and enforces its
budget. Bounded checks and clean scans remain evidence rather than correctness
proofs or security guarantees.
Section~\ref{sec:method} formalizes these distinctions through the acceptance
ladder.

%% file: sections/method.tex
\vspace{-0.5em}
\section{Proposed Framework}
\label{sec:method}

\begin{figure}[t]
\centering
\includegraphics[width=0.9\columnwidth]{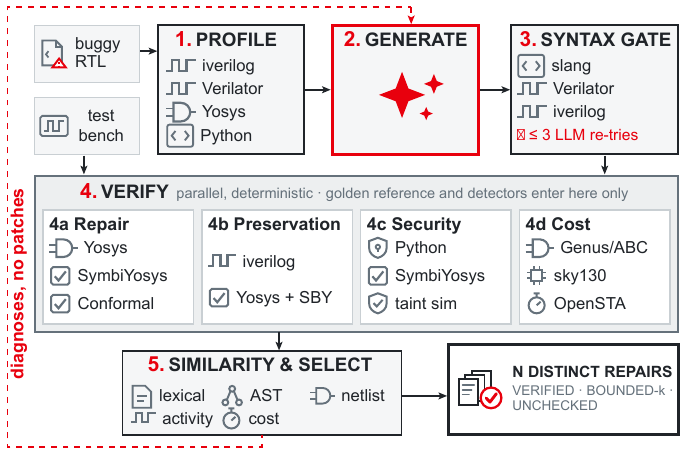}
\caption{The \system{} repair flow. Profiling and failure localization define
generation directions; independent repair, preservation, security, and cost
checks determine acceptance; failure evidence guides later rounds; and
similarity selection retains distinct verified variants.}
\label{fig:pipeline}
\vspace{-3pt}
\end{figure}

\system{} uses a generate--check--refine loop. Here, a bug is an RTL error that
violates specified functional or security behavior; the scope covers syntax
faults, functional control or datapath errors, and rule-based CWE weaknesses.
The LLM generates complete variants, design checks reject unsupported repairs,
and failure evidence guides the next round. This separation lets the model
explore repairs without allowing it to decide acceptance.

\subsection{Repair Goal and Acceptance Evidence}
\label{sec:method-overview}
Given the buggy design $D_b$, testbench $T$, and required checks $O$, \system{}
generates variants and runs $O$ under fixed tool configurations. Fixed inputs
and configurations make acceptance repeatable and independent of LLM judgment.
\emph{Acceptance} means selection after every required check passes. Functional
benchmarks may provide a golden design $D_g$; security benchmarks provide
weakness detectors. Neither $D_g$ nor a ground-truth repair enters the prompt.
The framework supports any LLM backend and human- or machine-written $D_b$. Each variant receives a record containing syntax, simulation,
formal, security, and synthesis results, together with proof depth, tool status,
and measured cost.
Functional verdicts are ordered $\V{}>\B{}\text{-}k>\U{}$ and security verdicts
as clean $>$ clean-bounded $>$ unchecked, with \F{} below both. An unavailable
optional check remains \U{}; an unavailable required check blocks selection.
A selected variant is an \emph{evidence-complete repair} only when repair and
preservation are checked.

The record additionally provides feedback for subsequent generation rounds.
A candidate receives the score of its
highest completed verification stage:
\vspace{-0.6em}
\begin{equation}
\begin{gathered}
r(v)\in\{0,.1,.3,.5,.6,.8,.9,1\},\\[-2pt]
\operatorname{stage}(v)\in
\{\mathrm{fail,syn,sim,B,V,pres,sec,full}\}.
\end{gathered}
\label{eq:rung}
\end{equation}
\vspace{-0.3em}
Each value is a predetermined credit for that stage, increasing along the
checking order so later rounds can prefer a direction that advanced further.

\subsection{Profiling and Structured Variant Generation}
\label{sec:method-profile}
Repeated prompts can return the same edit, so \system{} structures generation
using a profile of clocks, resets, counters, FSMs, memories, arithmetic,
handshakes, hierarchy, and security assets. The profile enables only applicable
transformations.
These choices are grouped into five axes: architecture, control,
datapath, coding style, and security hardening. An inapplicable axis becomes
$\{\textit{none}\}$ with a recorded reason. One value from each applicable axis
forms $z=(z_{\rm arch},z_{\rm ctrl},z_{\rm dp},z_{\rm style},z_H)$. The vector
directs the model to produce a complete repair using a stated change class; it
is not an RTL modification. Candidates share the bug evidence and output
contract but explore different strategies. The grid
$\mathcal G=\prod_a\mathcal M_a$ contains all directions
allowed by the projected menus $\mathcal M_a$. The first round covers this grid before exploiting any one strategy. Starting
from a canonical vector, the next vector maximizes its minimum Hamming distance
from the vectors already issued:
\setlength{\belowdisplayskip}{1pt} \setlength{\belowdisplayshortskip}{1pt}
\setlength{\abovedisplayskip}{1pt} \setlength{\abovedisplayshortskip}{1pt}
\begin{equation}
z^*=\arg\max_{z\in\mathcal G\setminus Z}
      \min_{u\in Z}\sum_a \mathbf 1[z_a\ne u_a].
\label{eq:direction}
\end{equation}
If $|\mathcal G|<N$, every vector is used before any is repeated, and the run
records that the design offers limited directions. Subsequent rounds adapt the search using observed verification rewards.
Each option starts with $\mathrm{Beta}(1,1)$; reward $r$ updates
$\alpha\leftarrow\alpha+r$ and $\beta\leftarrow\beta+1-r$. Thompson
sampling~\cite{thompson1933} is used because it balances choices that produced
strong evidence with underexplored choices when few trials are available.
Resampling avoids an observed vector when an unused one exists.

\begin{figure}[!t]
\centering
\scalebox{0.85}{\begin{tcolorbox}[enhanced,colback=white,colframe=cInk,boxrule=0.7pt,
 arc=1mm,left=4pt,right=4pt,top=2pt,bottom=2pt,before skip=0pt,after skip=3pt,
 title={Prompt anatomy: three configurations},
 colbacktitle=cCrim,coltitle=white,fonttitle=\bfseries]
\textbf{BARE \textcolor{cGray}{| minimal request}}\ \ \emph{The Verilog
module below has a bug. Find it and fix it.} \texttt{[numbered buggy RTL]}\par
\smallskip
\textbf{UNGUIDED \textcolor{cGray}{| add context}}\ \
\texttt{[structure and security assets]} \texttt{[applicable
transformations]} \texttt{[ranked regions and failing trace]}
\texttt{[candidate transformation direction]}\par
\smallskip
{\color{cCrim}\textbf{GUIDED | add diagnosis}}\ \
\texttt{[prior attempt's blocking gate]} \texttt{[region;
expected/observed behavior]} \texttt{[counterexample, if available]}
\texttt{[repair hypothesis; repeated failure]}
\tcblower
\textbf{Shared output contract.} Return one complete, self-contained
synthesizable Verilog module preserving the interface: neither a patch
fragment nor explanatory prose.
\end{tcolorbox}}
\caption{Prompt structure used to generate repair variants. Fixed text is
verbatim; brackets identify data filled and retained by the framework.}
\label{fig:prompt}
\end{figure}

Generation also uses failure location. Replay and switching coverage assign
each source region the Ochiai score~\cite{abreu2007accuracy}
$s(e)=f_e/\sqrt{F(f_e+p_e)}$, where $f_e$ and $p_e$ count failing and passing
checks that exercise region $e$, and $F$ is the number of failing checks. A
flat ranking triggers statement-level analysis. The prompt then receives the
ranked region, its signal cone, and a short failing trace. These signals localize the observed failure region but are not interpreted as identifying a minimal triggering condition.

Proposed \system{} has three configurations. \emph{Bare} receives the source
and repair request; \emph{unguided} adds the profile, failure evidence, and
transformation vector; and \emph{guided} adds the prior candidate's earliest
failed gate, counterexample, and repair hypothesis. Two repeated failure
signatures cause the guidance to revise that hypothesis. Each configuration returns one complete module. The framework
rejects malformed outputs and permits three syntax-correction iterations using
tool diagnostics. Algorithm~\ref{alg:round} summarizes the same loop: profile and
localize, generate, check, select, diagnose, and repeat.

\begin{algorithm}[!t]
\caption{Verification-guided RTL variant generation}
\label{alg:round}
\footnotesize
\begin{algorithmic}[1]
\Require buggy RTL $D_b$; testbench $T$; required checks $O$; prompt configuration $p$;
 target $N$; round limit $R$; similarity limit $\tau$
\Ensure selected repairs $A$ with their recorded evidence
\State $(P,L)\gets\textsc{ProfileAndLocalize}(D_b,T)$
\State $\mathcal G\gets\prod_a\mathcal M_a(P)$; initialize each option as
 $\mathrm{Beta}(1,1)$
\State pool $C\gets\emptyset$; selected set $A\gets\emptyset$; diagnoses
 $\Delta\gets\emptyset$
\For{$q=1$ \textbf{to} $R$}
 \State $m\gets N-|A|$ \Comment{generate only the missing variants}
 \If{$p$ is bare} $Z_q\gets m$ copies of the empty direction
 \ElsIf{$q=1$} $Z_q\gets m$ max-min directions from $\mathcal G$
 \Else{} $Z_q\gets m$ Thompson-sampled, nonrepeated directions
 \EndIf
 \ForAll{$z\in Z_q$}
  \State $v\gets\textsc{Generate}(D_b,P,L,z,\Delta,p)$
  \State extract one complete module; run at most three syntax repairs
  \State $E_v\gets\textsc{Check}(v,T,O)$; $C\gets C\cup\{(v,E_v,z)\}$
  \State update the options in $z$ using $r(v)$ from Eq.~\eqref{eq:rung}
 \EndFor
 \State $(A,d_{\min})\gets\textsc{SelectDiverse}(C,A,N,\tau)$
 \If{$p$ is guided} $\Delta\gets\textsc{Diagnose}$ earliest failed gates
 \EndIf
 \State \textbf{stop} on target, measured hardware collapse, two stalled rounds,
 round limit, or budget
\EndFor
\State \Return $A$ with evidence, depths, tool flags, and stop reason
\end{algorithmic}
\end{algorithm}

\subsection{Repair, Preservation, and Security Checks}
\label{sec:method-verify}
Four independent checks cover repair, preservation, security,
and implementation cost. Separate verdicts prevent one successful check from
hiding a failed or unevaluated result.

\textbf{Repair.} 
The repair check determines whether the candidate eliminates the targeted functional defect. 
A testbench failure rejects it outright. Otherwise, bounded model
checking~\cite{biere1999bmc} builds a miter, a construction that searches for
an input $x$ on which the candidate $v$ and the golden design $g$ disagree by depth
$k$:
\begin{equation}
\exists\,x,\ \exists\,t\le k:\quad y_v(x,t)\ \ne\ y_g(x,t).
\label{eq:miter}
\end{equation}
A solution returns a counterexample and yields \F{}; no solution excludes
disagreement only within the encoded scope. Checking starts at depth 8 and then
uses $k=\min(64,2w+d)$, where $w$ is the widest counter or FSM register and $d$
is pipeline depth. The $2w+d$ term scales the checking horizon with sequential-state width and pipeline depth, while the upper bound limits solver cost.

No mismatch yields
\V{} for a state-free miter and \B{}-$k$ otherwise. This schedule is a cost
heuristic, not a completeness claim; commercial equivalence results are not
part of the reported verdicts.

\textbf{Preservation.} The preservation check evaluates whether defect removal modifies behavior that was previously correct in the original design.
A mismatch rejects preservation, while agreement is evidence for those rows alone. 
If all observations fail, preservation is marked as not applicable rather than treating an empty comparison set as successful evidence.

\textbf{Security:} Security rules and assertion templates are derived from MITRE
hardware CWE definitions~\cite{mitre2025mihw} and instantiated with assets found
during profiling or declared by the benchmark. Static AST rules check missing
security-state resets, debug bypasses, and overlapping address ranges. Formal
assertions check write-after-lock and reset behavior, while shadow-netlist
information-flow tracking checks explicit and implicit flows from secret or
privileged state to observable outputs~\cite{hu2021ift}. Before-and-after
findings are \emph{fixed}, \emph{persisted}, or \emph{regressed}; vacuity checks
reject fixes that only make a guard unreachable. A persisted target, regression,
lost asset, or masked fix rejects the candidate.

\textbf{Implementation cost.} The fourth check measures area, delay, and power
by synthesis under shared stimulus. Each metric must stay within
$(1+\varepsilon)$ of its reference, with $\varepsilon=0.20$ here, and below 50
cells, an eight-cell allowance replaces a percentage bound that a small baseline
would destabilize. A metric without a valid reference remains \U{} and cannot
support any overhead claim.

\setlength{\textfloatsep}{0pt}
\begin{algorithm}[!t]
\caption{Evidence-first diversity selection}
\label{alg:similarity}
\footnotesize
\begin{algorithmic}[1]
\Require pool $C$; selected $A$; obligations $O$; target $N$; threshold
 $\tau$; view weights $w$
\Ensure updated $A$, dispersion $d_{\min}$
\State $E\gets\{v\in C:v\text{ passes every required check in }O\}$
 \Comment{evidence first}
\State \textbf{if} $E=\emptyset$ \textbf{then return} $(A,1)$
\State score each eligible pair on the views $V$ it supports
 \Comment{Eqs.~\eqref{eq:source-sim}--\eqref{eq:act-dist}}
\If{$|E|\ge3$}\ drop any view that separates nothing, agreeing within
 $10^{-3}$ on most pairs
\EndIf
\State renormalize retained $w$ and fuse:
 $S(i,j)\gets\sum_{u\in V}w_uS_u(i,j)$
\State keep earlier selections; if $A=\emptyset$, seed the most isolated
 member of $E$
\While{$|A|<N$ \textbf{and} $E\setminus A\ne\emptyset$}
 \State $i^{*}\gets\arg\max_{i\in E\setminus A}\min_{j\in A}(1-S(i,j))$,
  the eligible candidate farthest from $A$
  \Comment{Eq.~\eqref{eq:fusion}}
 \State \textbf{if} $\max_{j\in A}S(i^{*},j)>\tau$ \textbf{then break}
  \Comment{too close to a selection}
 \State \textbf{else} $A\gets A\cup\{i^{*}\}$
\EndWhile
\State $d_{\min}\gets\min_{i\ne j\in A}(1-S(i,j))$, or $1$ if $|A|<2$
\State \Return $(A,d_{\min})$ with retained views and drop reasons
\end{algorithmic}
\end{algorithm}

\vspace{-1em}
\subsection{Similarity Views and Selection}
\label{sec:similarity}
Verification determines candidate acceptability but does not establish whether two variants correspond to distinct repair implementations.
Selection, therefore, estimates similarity from five
views in $[0,1]$, where 1 denotes indistinguishable. These views organize verified
variants; they do not strengthen a repair verdict.

\textbf{Source views.} Lexical fingerprints $H_i$ are built from five-token
BLAKE2b hashes by four-hash winnowing~\cite{schleimer2003winnowing}. The AST
view discards identifiers and literals, then forms pq-gram bag $Q_i$ from two
ancestor nodes and three adjacent child nodes~\cite{augsten2010pq}. The two similarities are
\begin{equation}
\sm{lex}(i,j)=\frac{|H_i\cap H_j|}{|H_i\cup H_j|},\qquad
\sm{ast}(i,j)=\frac{2\,|Q_i\cap Q_j|}{|Q_i|+|Q_j|}.
\label{eq:source-sim}
\end{equation}
Thus, renaming leaves $\sm{ast}$ unchanged, while restructuring lowers
it. A parse failure makes this unavailable for the affected pair.

\textbf{Netlist view.} Source-level differences may be eliminated after synthesis.
Technology-independent synthesis~\cite{wolf2016yosys} therefore produces a
labeled cell-port graph. Three Weisfeiler--Lehman rounds encode local wiring in
$\phi_i$~\cite{shervashidze2011wl}, while normalized Laplacian eigenvalues
$\Lambda_i$ encode global structure:
\vspace{-4pt}
\begin{equation}
\sm{net}(i,j)=\tfrac12\big[\cossim(\phi_i,\phi_j)+e^{-W(\Lambda_i,\Lambda_j)/\sigma}\big].
\label{eq:net-sim}
\end{equation}
Here $W$ is earth-mover distance. Spectra are exact below 2{,}000 nodes and
approximated for larger graphs.

\textbf{Activity and cost views.} Shared stimulus produces a 32-bin toggle-rate
histogram $h_i$. Activity distance combines its earth-mover distance with an
energy proxy:
{\small
\begin{equation}
d_E(i,j)=W(h_i,h_j)+0.2\,|E_i-E_j|,\quad
E_i=\textstyle\sum_n \alpha_n\,\mathrm{fanout}_n,
\label{eq:act-dist}
\end{equation}}
where fanout approximates capacitance. The transform $e^{-d_E/\sigma}$ uses the
pool's median positive distance $\sigma$. The cost view applies the same
transform to available, budget-normalized PPA metrics. Since $E_i$ is not
normalized, it may dominate the histogram term in large netlists.

\textbf{Combining the views.} Lexical, AST, netlist, activity, and cost views
use weights $0.10$, $0.20$, $0.30$, $0.30$, and $0.10$. Netlist and activity
receive $0.30$ because they measure synthesized structure and behavior; AST
receives $0.20$ for source structure. Lexical and cost receive $0.10$ because
lexical variation can arise without structural modification, whereas PPA similarity does not directly characterize functional structure.
These heuristic
weights were fixed before the campaign and not tuned. With at least three
variants, a view is dropped if more than
half of its pair scores agree within $10^{-3}$; the remaining weights are
renormalized into $S(i,j)$. Selection starts with the most isolated feasible
variant and repeatedly chooses
\begin{equation}
\max_{i\notin A}\ \min_{j\in A}\ \big(1-S(i,j)\big).
\label{eq:fusion}
\end{equation}
It admits the variant only if its similarity to every member of $A$ is at most
$\tau$. The archive is monotone across rounds, even if later pool-dependent
rescaling changes an old score. Target attainment therefore uses the final
minimum distance $d_{\min}$, not archive size alone. 
If a hardware-level view is excluded, source-distinct candidates may remain functionally or structurally equivalent after synthesis.
Hence, Section~\ref{sec:results} uses exact equivalence to
audit identical-cost groups; the proxy itself is not proof of hardware
diversity.

%% file: sections/setup.tex
\vspace{-1em}
\section{Experimental Setup}
\begin{table}[!t]
\centering
\caption{Major tools used in the evaluation.}
\scalebox{0.85}{\begin{minipage}{\columnwidth}
\centering
\scriptsize
\setlength{\tabcolsep}{2.5pt}\renewcommand{\arraystretch}{0.92}
\begin{tabular}{@{}p{0.31\columnwidth}p{0.61\columnwidth}@{}}
\rowcolor{cCrim}
\thd{Tool} & \thd{Purpose}\\
Icarus Verilog & Simulation, testbench replay, and VCD traces.\\
\rowcolor{cCrim!5}
Slang/Verilator & SystemVerilog syntax and elaboration checks.\\
Yosys & Open-source synthesis and miter construction.\\
\rowcolor{cCrim!5}
SymbiYosys & Bounded formal verification of RTL properties.\\
Cadence Genus & Commercial synthesis and PPA measurement.\\
\rowcolor{cCrim!5}
Cadence Conformal & Commercial logic-equivalence checking.\\
OpenSTA & Open-source static timing analysis.\\
\bottomrule
\end{tabular}
\end{minipage}}
\vspace{-2pt}
\end{table}

\label{sec:setup}
\vspace{-0.2em}
The primary campaign crosses three prompt configurations, three backends, and 32
benchmarks, giving 288 runs (Table~\ref{tab:results}(a)). A subsequent analysis
adds 20 guided-\Codex{} input-transformation controls, for 308 manifests and 629 selected
variants. Every run uses seed zero and a nominal 20\% cost budget; each round
requests only the missing candidates. \Codex{} (GPT-5.5), \Claude{} Sonnet in print
mode, and the \Grok{}-4.6 CLI without tool access span three model families and
two interface styles. Cadence Genus synthesizes each variant and reports cell
area and count, worst-path delay, and power.

\begin{table*}[!t]
\centering
\caption{Primary campaign and outcomes: (a) suite limits; (b) CWE configured
selection; (c)
Hack@DAC stages; (d) CirFix resource use; (e) depth of bounded verdicts.}
\label{tab:results}
\scalebox{0.90}{\begin{minipage}{\textwidth}
\centering
\scriptsize
\tablecompact\setlength{\tabcolsep}{1.5pt}\renewcommand{\arraystretch}{0.82}
\begin{minipage}[t]{0.265\textwidth}
\centering
\textbf{(a) Campaign limits}\\[1pt]
\begin{tabular}{lrrrr}
\rowcolor{cCrim}
\thd{Suite} & \thd{Cases} & \thd{$N$} & \thd{$\tau$} & \thd{Rnds}\\
CirFix & 21 & 5 & 0.70$^\dagger$ & 3\\
\rowcolor{cCrim!5}
CWE & 10 & 3 & 0.85 & 1\\
Hack@DAC & 1 & 5 & 0.70 & 3\\
\bottomrule
\end{tabular}

\vspace{4pt}
\textbf{(b) CWE configured selection}\\[1pt]
\begin{tabular}{lrrr}
\rowcolor{cCrim}
\thd{Prompt} & \thd{Sec. set} & \thd{Target} & \thd{Clean / failed}\\
Bare & 26/27 & 1/27 & 76 / 5\\
\rowcolor{cCrim!5}
Unguided & 27/27 & 15/27 & 65 / 15\\
Guided & 27/27 & 16/27 & 68 / 12\\
\bottomrule
\end{tabular}
\end{minipage}\hfill
\begin{minipage}[t]{0.265\textwidth}
\centering
\textbf{(c) Hack@DAC candidate outcomes}\\[1pt]
\begin{tabular}{llrrrr}
\rowcolor{cCrim}
\thd{Backend} & \thd{Prompt} & \thd{G} & \thd{S} & \thd{R} & \thd{D}\\
Codex & Bare & 11 & 11 & 8 & 2\\
 & Unguided & 7 & 7 & 7 & 4\\
 & Guided & 8 & 8 & 8 & 5\\
\rowcolor{cCrim!5}
Sonnet & Bare & 13 & 13 & 5 & 1\\
\rowcolor{cCrim!5}
 & Unguided & 9 & 7 & 7 & 3\\
\rowcolor{cCrim!5}
 & Guided & 11 & 10 & 9 & 5\\
Grok & Bare & 13 & 6 & 3 & 1\\
 & Unguided & 5 & 5 & 5 & 5\\
 & Guided & 5 & 5 & 5 & 5\\
\bottomrule
\end{tabular}
\end{minipage}\hfill
\begin{minipage}[t]{0.285\textwidth}
\centering
\textbf{(d) CirFix resource use}\\[1pt]
\begin{tabular}{llrrr}
\rowcolor{cCrim}
\thd{Backend} & \thd{Prompt} & \thd{Tok.\ (M)} & \thd{Time (h)} & \thd{Deliv.}\\
Codex & Bare & 0.775 & 3.1 & 12/21\\
 & Unguided & 1.750 & 4.0 & 18/21\\
 & Guided & 1.849 & 4.1 & 20/21\\
\rowcolor{cCrim!5}
Sonnet & Bare & 0.335 & 1.7 & 6/21\\
\rowcolor{cCrim!5}
 & Unguided & 0.851 & 3.0 & 13/21\\
\rowcolor{cCrim!5}
 & Guided & 0.983 & 3.6 & 13/21\\
Grok & Bare & 3.568 & 5.5 & 14/21\\
 & Unguided & 3.477 & 10.6 & 14/21\\
 & Guided & 4.055 & 11.4 & 18/21\\
\bottomrule
\end{tabular}
\end{minipage}\hfill
\begin{minipage}[t]{0.18\textwidth}
\centering
\textbf{(e) Depth $k$ of bounded verdicts}\\[1pt]
\begin{tabular}{@{}lrr@{}}
\rowcolor{cCrim}
\thd{Range of $k$} & \thd{Repair} & \thd{Preserv.}\\
$k\le 9$ & 98 & 31\\
\rowcolor{cCrim!5}
$10\le k\le 31$ & 192 & 278\\
$k\ge 32$ & 0 & 121\\
\bottomrule
\end{tabular}
\end{minipage}
\par\vspace{1pt}
{\scriptsize\raggedright(a) Cases is the number of benchmark designs, $N$ is the target
repair count, $\tau$ is the similarity limit, and Rnds is the round limit;
$^\dagger$Sonnet CirFix manifests use 0.85, while
Codex/Grok use 0.70. CWE ran ten cases but detector conclusions use nine.
(b) Security-clean selection; functional repair is evaluated separately.
(c) G generated, S syntax, R repair verified, D delivered; column totals are
82, 72, 57, and 31, matching the pooled manifests. (d) tokens in millions,
time in hours. (e) counts by bounded formal depth.\par}
\end{minipage}}
\vspace{-5pt}
\end{table*}

\textbf{Benchmarks.} The three suites exercise complementary repair settings.
We use 21 seeded RTL bugs from the published CirFix
benchmark~\cite{ahmad2022cirfix}. We created the CWE suite for security-repair testing
from ten hand-written RTL examples based on MITRE hardware
weaknesses~\cite{mitre2025mihw}. The detector covers nine examples; the
missing-default FSM remains in the 288-run total but is excluded from security
conclusions. For Hack@DAC, we used a buggy OpenTitan benchmark. We extracted the affected register-arbiter logic,
replaced package-specific types, made one equivalent operator rewrite, and
added a focused testbench. The LLM repairs only the affected module. Across the
three suites, inputs are human-written or human-seeded, and neither golden RTL
nor a ground-truth repair enters the prompt. 
The interface can also accept machine-generated RTL.

\textbf{Controls.} Within a suite, the prompt configurations share tools,
oracles, and target sizes; only the prompt evidence changes
(Fig.~\ref{fig:prompt}). One LLM call means sending one prompt to a backend and
receiving one response. Because adaptive stopping changes the number of calls
and tokens, the experiment compares complete configurations rather than equal
compute. CWE uses one generation round, so it cannot measure whether feedback
helps in a later round. We added 20 guided \Codex{} runs to test sensitivity to
surface changes: ten rename identifiers and ten alter RTL structure. We also
reselected stored candidates at $\tau\in\{0.70,0.85,1.0\}$ to measure the
similarity-threshold tradeoff. Reselection uses no new LLM calls. We did not
repeat the campaign with different similarity weights.

\textbf{Reproducibility.} Each run retains its resolved configuration,
rendered knowledge pack, generated RTL, per-tool logs, verdicts, tokens, and
wall-clock time. One anonymized design example includes rendered prompts,
delivered variants, parameters, and evidence.
Potential hallucinations, such as incomplete modules or
unsupported constructs, remain in the generated RTL and logs and are rejected
when extraction or syntax checking fails. The logs identify each model family, but they do not contain an exact model
snapshot or the provider's default decoding temperature. We therefore report
these settings as unavailable.

\textbf{Tools.} The flow supports both open-source and commercial EDA tools for
syntax checking, simulation, formal verification, synthesis, and PPA analysis.
The major tools used in this evaluation are listed below. Equivalent tools can
be integrated through the same stage interfaces.

\textbf{Metrics.} For $n$ runs, selected count $a_r$, and final dispersion
$d_r$, configured delivery $D$ and target attainment $H$ are
\[
D=\frac1n\sum_r\mathbf 1[a_r\ge1],\qquad
H=\frac1n\sum_r\mathbf 1[a_r=N\ \land\ d_r\ge1-\tau].
\]
These campaign metrics are reported alongside, never instead of, the typed
evidence distribution. The 63 CirFix observations
per prompt configuration represent 21 bugs on three backends rather than 63 independent bugs, and with one seed per configuration every rate reported below is
descriptive rather than inferential.

\vspace{-0.7em}

%% file: sections/results.tex
\section{Results}
\label{sec:results}
We have evaluated the variants in three steps. It first measures
whether structural context and iterative feedback increase checked repair
delivery. Next, it examines the verification evidence for each selected
variant. Finally, it compares the RTL and synthesized implementations to
determine whether generated variants are distinct.

\subsection{\texorpdfstring{\RQOne{}}{RQ1}: Checked Repair Delivery with Context and Feedback}
\label{sec:res-guidance}
\begin{figure}[t]
\centering
\includegraphics[width=0.7\columnwidth]{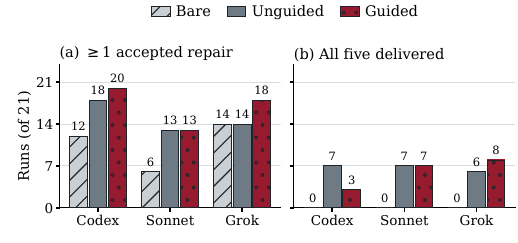}
\vspace{-1em}
\caption{CirFix repair outcomes by backend and prompt configuration. (a)
Runs producing at least one checked repair; (b) runs reaching the full
five-candidate target. Context and feedback increase checked
repair delivery, although five-variant target attainment is not monotonic.}
\label{fig:guidance}
\end{figure}

\RQOne{} evaluates whether design context and verification feedback help an LLM
produce at least one checked repair. On CirFix, delivery increases from 32/63
runs with bare prompting to 45/63 with structural context and 51/63 with context
plus diagnosis (Fig.~\ref{fig:guidance}). Context identifies relevant RTL
structures, while diagnosis reports the check that prevented acceptance in the
previous round. These inputs therefore reduce repeated or unsupported edits.

The backend results show that the benefit is configuration dependent. Diagnosis
increases solved benchmarks from 18 to 20 for \Codex{} and from 14 to 18 for
\Grok{}, while \Sonnet{} remains at 13 of 21. Guided CirFix also uses 47\% more
new tokens than bare prompting because adaptive rounds continue when evidence is
missing. Consequently, the results establish an association under the evaluated
budgets rather than a compute-matched causal effect. The reduction in runs
without a checked repair provides hardware designers with more usable
candidates and gives future RTL-repair studies a reproducible comparison point.

\vspace{-1em}
\subsection{\texorpdfstring{\RQTwo{}}{RQ2}: Verification Evidence for Selected Repairs}
\label{sec:res-evidence}

\RQTwo{} determines whether each selected variant has sufficient evidence to
support a repair claim. All 430 functional selections have checked repair and
preservation results. Among them, 140 reach an unbounded \V{} repair verdict,
while 290 reach \B{}-$k$ at $k\le 20$; preservation reaches $k\ge 32$ for 121
variants (Table~\ref{tab:results}(e)). Thus, a common delivery count can contain
different proof strengths. PPA results provide a second distinction: 236
CirFix variants satisfy the budget, 81 exceed it, and 82 lack a complete cost
result. Hack@DAC has measured PPA but no valid reference report, so an overhead
claim remains \U{}. A hardware designer can therefore select a variant according
to the evidence required by the design rather than according to file count.

The CWE results further show why functional and security evidence must remain
separate. All 199 selected candidates satisfy the configured differential
security check; however, the functional replay does not establish repair or
preservation. Consequently, \system{} records their functional status as \U{}
and does not classify them as functionally checked repairs. The original
selection rates of 26/27, 27/27, and 27/27 do not express this difference.

Table~\ref{tab:evidence}(b) explains how stronger checks change the supported
claim. Compilation retains 629 candidates across 44 benchmarks, functional
replay retains 430 across 21, unbounded equivalence retains 140 across five,
and the cost gate retains 109 across four. Therefore, functional replay alone
reports 3.9$\times$ more repairs than the full set of applicable checks. The
remaining entries are \U{} rather than \F{} when a check is unavailable or only
a bounded result is present. This progression provides a consistent basis for
comparing repair systems that report evidence at different depths.

The input-transformation results examine whether the security outcome depends
on the exact RTL text. In the 30-run guided-\Codex{} study, a security-clean
candidate appears for 9/10 original designs and 7/10 designs after either
identifier renaming or structural perturbation. Across 90 rescans, the
detectors record 53 fixed, 70 persistent, and 8 regressed findings, with no
masked finding; one scan may contain multiple findings. The 70\% retained rate
after each transformation shows that exact-text recall is not the only source
of successful repairs. However, the small sample and the uncovered
missing-default FSM limit the strength and scope of this result.
\vspace{-.9em}

\begin{table}[!t]
\centering
\caption{Verification outcomes for 629 selected variants across 308 runs: (a)
verification outcomes; (b) variants retained as the required checks are applied.}
\label{tab:evidence}
\scalebox{0.8}{\begin{minipage}{\columnwidth}
\centering
\scriptsize
\setlength{\tabcolsep}{1.4pt}\renewcommand{\arraystretch}{0.92}
\begin{minipage}[t]{0.465\columnwidth}
\vspace{0pt}
\centering
\textbf{(a) Verification outcomes}\\[2pt]
\begin{tabular}{@{}lrrr@{}}
\rowcolor{cCrim}
\thd{Check} & \thd{CirFix} & \thd{CWE} & \thd{H@D}\\
Variants $n$ & 399 & 199 & 31\\
\rowcolor{cCrim!5}
Runs deliv. & 128/189 & 94/110 & 9/9\\
Repair V/B/U & 109/290/0 & 0/0/199 & 31/0/0\\
\rowcolor{cCrim!5}
Pres. B/U & 399/0 & 0/199 & 31/0\\
PPA P/F/U & 236/81/82 & 0/0/199 & 0/0/31\\
\rowcolor{cCrim!5}
Sec. C/CB & --/-- & 63/136 & --/--\\
\bottomrule
\end{tabular}

\end{minipage}\hfill%
\begin{minipage}[t]{0.515\columnwidth}
\vspace{0pt}
\centering
\textbf{(b) Criteria applied in sequence}\\[2pt]
\begin{tabular}{@{}lrrrr@{}}
\rowcolor{cCrim}
\thd{\shortstack[l]{Criterion}} & \thd{CirFix} & \thd{CWE} &
\thd{H@D} & \thd{Bugs}\\
\shortstack[l]{Compiles, parses} & 399 & 199 & 31 & 44\\
\rowcolor{cCrim!5}
\shortstack[l]{$+$ testbench} & 399 & 0$^{\dagger}$ & 31 & 21\\
\shortstack[l]{$+$ preservation} & 399 & 0$^{\dagger}$ & 31 & 21\\
\rowcolor{cCrim!5}
\shortstack[l]{$+$ equivalence} & 109 & 0$^{\dagger}$ & 31 & 5\\
\shortstack[l]{$+$ security scan} & 109 & 0$^{\dagger}$ & 31 & 5\\
\rowcolor{cCrim!5}
\shortstack[l]{$+$ cost gate} & 109 & 0$^{\dagger}$ & 0$^{\dagger}$ & 4\\
\bottomrule
\end{tabular}
\end{minipage}
\par\vspace{2pt}
\begin{minipage}{\columnwidth}\scriptsize
\setlength{\parindent}{0pt}
$^{\dagger}$Check not evaluable on that corpus, not failed. Each row also
represents its check class: testbench (CirFix, MEIC, R3A), cost gate
(ChipSeek, LLM-VeriPPA).
\end{minipage}
\end{minipage}}
\end{table}

\subsection{\texorpdfstring{\RQThree{}}{RQ3}: Hardware Equivalence of Generated Variants}
\label{sec:res-diversity}

\RQThree{} evaluates whether source-level variation produces distinct hardware
implementations. Bare prompting never reaches the five-candidate CirFix target,
while unguided and guided prompting reach it in 20/63 and 18/63 runs. For
\Codex{}, diagnosis increases checked delivery from 18 to 20 benchmarks but
reduces target attainment from seven to three. Verification feedback helps the
model correct a rejected edit, but it can also direct later variants toward the
same repair. Unequal compute and the \Sonnet{} threshold discrepancy prevent a
causal interpretation.

\begin{figure}[t]
\centering
\includegraphics[width=0.75\columnwidth]{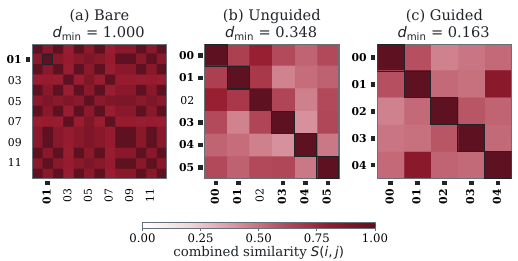}
\caption{Combined similarity score $S(i,j)$ for one matched CirFix/\Codex{} case;
bold labels and boxed cells identify selected candidates. The
lower $d_{\min}$ under guided prompting shows that verification feedback can
reduce dispersion among selected variants.}
\label{fig:matrices}

\end{figure}

The matched case in Fig.~\ref{fig:matrices} illustrates this tradeoff.
Unguided context selects five candidates with final $d_{\min}=0.348$, whereas
guided selection ends at 0.163 because an archived pair is rescored above
$\tau$. A deterministic threshold sweep confirms the expected tradeoff:
raising $\tau$ from 0.70 to 0.85 to 1.0 increases the mean selected fraction from
78.9\% to 87.1\% to 99.3\%, while the mean $d_{\min}$ falls from 0.653 to 0.570 to
0.426. Thus, target size and dispersion are competing operating points rather
than similar success measures.

Selection diversity is calculated as
$d_{\min}=\min_{i\ne j}(1-S(i,j))$; a larger value means that the closest pair
is less similar. Because this score combines source and hardware views, exact
equivalence determines whether the synthesized variants differ functionally.
A Yosys SAT miter
proves all 408 variant pairs from 16
identical-PPA groups functionally equivalent, with no counterexample or timeout.
In contrast, the netlist view places only 192/372 checkable pairs above 0.999
similarity. The combined score therefore measures representation-level
variation, not guaranteed implementation diversity. Hardware designers and
future repair tools should group final candidates by exact equivalence and use
the less expensive similarity views only to prioritize comparisons.

\subsection{Resource Cost}
\label{sec:res-hackdac}
Token and wall-clock costs are important when selecting an iterative-repair
backend. Table~\ref{tab:results}(d) shows that \Codex{} and \Sonnet{} reuse
cached prompt context, whereas the \Grok{} CLI reprocesses the full input for
each candidate and feedback round. Consequently, \Grok{} uses 3.477--4.055
million new tokens and 5.5--11.4 hours, while the cached interfaces complete in
1.7--4.1 hours. Caching therefore favors \Codex{} and \Sonnet{} for repeated
calls in this workflow; these values describe interfaces, not model capability.
\vspace{-0.5em}

%% file: sections/closing.tex
\section{Discussion and Limitations}
\label{sec:limits}
\subsection{Verification Evidence for RTL Repair}
\system{} applies the same checks to human-written and machine-generated RTL.
Acceptance requires bug removal, off-trigger preservation, security status, and
PPA feasibility when a reference is available. Recorded counterexamples, proof
depths, and PPA results allow hardware designers to select evidence appropriate
to each design. All 408 variant pairs within identical-PPA groups implement the same function, so final
variants should be grouped by exact equivalence. The 199 security-selected CWE
candidates lack checked functional replay; their security results therefore do
not establish a functional repair.
\vspace{-0.8em}
\subsection{Limitations}
\textbf{Security.} The current repair scope covers syntax faults, functional
control or datapath bugs evaluated with a testbench or golden design, and known
MITRE CWE weaknesses expressible through configured rules, assertions, and
information-flow checks. \system{} may not detect a new vulnerability
outside these definitions. Future work should detect and localize unknown
vulnerabilities and return them to the repair loop.

\textbf{Runtime and Token Cost.}
Repeated CLI requests consume substantial time and tokens. Because the evaluated
\Grok{} CLI lacks prompt caching, each round reprocesses fixed context. Future
implementations should cache this prefix, transmit only new feedback, and reuse
verification artifacts.

\vspace{-0.5em}
\section{Conclusion}
\label{sec:conclusion}
LLM-generated RTL repairs require evidence of bug removal, behavioral
preservation, security, and implementation cost. \system{} combines structured
variant generation with simulation, formal verification, security analysis,
synthesis, and evidence-based selection. The evaluation covers CirFix, MITRE CWE examples, and a Hack@DAC OpenTitan
bug across three backends and prompt configurations. Context and feedback
increase CirFix checked delivery from 32/63 to 51/63 runs. Among 629 selected
variants, 430 have checked functional evidence, while 199 CWE variants retain
only security results. Exact equivalence proves all 408 identical-PPA pairs
functionally identical, showing that RTL variation does not establish hardware
diversity. Counting equivalent variants as separate hardware
alternatives would overstate repair diversity and provide no independent
implementation choices. These results give researchers explicit evidence and diversity measures, while hardware designers receive candidates with inspectable verification status.
Future work includes detecting unknown weaknesses, using exact equivalence in
selection, and reducing iterative cost through caching and verification reuse.